\documentclass{article}
\usepackage{multirow}
\usepackage{graphicx} % Required for inserting images
\usepackage{dirtytalk}
\usepackage{authblk}
\usepackage{url}
\usepackage{xcolor}

\title{Whose LLM is it Anyway? Linguistic Comparison and LLM Attribution for GPT-3.5, GPT-4 and Bard}
\author{Ariel Rosenfeld$^{1,*}$, Teddy Lazebnik$^{2,3}$ \\ \(^1\) Department of Information Science, Bar Ilan University, Ramat Gan, Israel \\ \(^2\) Department of Information Systems, University of Haifa, Haifa, Israel \\ \(^3\) Department of Computing, Jonkoping University, Jonkoping, Sweden \\ \(*\) Corresponding author: ariel.rosenfeld@biu.ac.il }
\date{ }

\begin{document}

\maketitle

\begin{abstract}
    \noindent
    Large Language Models (LLMs) are capable of generating text that is similar to or surpasses human quality. However, it is unclear whether  LLMs tend to exhibit distinctive linguistic styles akin to how human authors do. Through a comprehensive linguistic analysis, we compare the vocabulary, Part-Of-Speech (POS) distribution, dependency distribution, and sentiment of texts generated by three of the most popular LLMS today (GPT-3.5, GPT-4, and Bard) to diverse inputs. The results point to significant linguistic variations which, in turn, enable us to attribute a given text to its LLM origin with a favorable 88\% accuracy using a simple off-the-shelf classification model. Theoretical and practical implications of this intriguing finding are discussed.  \\ \\
    
\end{abstract}

\section{Introduction}
Large Language Models (LLMs), such as GPT-3.5 \cite{ouyang2022training}, GPT-4 \cite{achiam2023gpt} and Bard \cite{bard}, have revolutionized and popularized natural language processing and AI, demonstrating human-like and super-human performance in a wide range of text-based tasks \cite{zhao2023survey}. While the layman may find the responses of LLMs hard to distinguish from human-generated ones \cite{layman_llm,new_intro_7}, a plethora of recent literature has shown that it is possible to successfully discern human-generated text from LLM-generated text using various computational techniques \cite{new_intro_llm_1,new_intro_llm_2,new_intro_llm_3}. Among the developed techniques, the \textit{linguistic approach}, which focuses on the structure, patterns, and nuances inherent in human language, stands out as a promising option that offers both high statistical performance \cite{herbold2023large} as well as theoretically-grounded explanatory power \cite{munoz2023contrasting}, as opposed to alternative \say{black-box} machine-learning techniques \cite{llm_math_1,llm_math_2}. 

Indeed, recent literature has shown that human and LLM-generated texts are, generally speaking, linguistically different across a wide variety of tasks and datasets including news reporting \cite{munoz2023contrasting}, hotel reviewing \cite{giorgi2023slept}, essay writing  \cite{herbold2023large} and scientific communication \cite{desaire2023distinguishing} to name a few. Common to these and similar studies is the observation that LLM-generated texts tend to be extensive and comprehensive, highly organized, follow a logical structure or formally stated, and present higher objectivity and lower prevalence of bias and harmful content compared to human-generated texts \cite{wu2023survey}. 

Extensive research into human-generated texts has consistently demonstrated the inherent diversity in human writing styles, resulting in distinct linguistic patterns, structures, and nuances \cite{liza,ws_general_3,ws_general_2}. Notably, highly successful techniques for author attribution \cite{ws_author_1,ws_author_2} and author profiling \cite{author_profile_1,author_profile_2} have leveraged linguistic markers to identify and differentiate between authors and their characteristics (sometimes referred to as Stylometrics \cite{stylometric}). These remarkable capabilities underscore both the richness and variability present in human-generated texts as well as the unique linguistic traits presented, rather consistently, by different authors. Unfortunately, to the best of our knowledge, a similar inquiry into LLM-generated texts has yet to take place. That is, it remains unclear whether different LLMs present distinct linguistic styles and, if so, could these linguistic markers be effectively used for LLM attribution (i.e., identifying which LLM has generated a given text).

In this work, we report on a comprehensive linguistic comparison of LLM-generated texts generated by three of the most popular LLMs today: GPT-3.5, GPT-4, and Bard. Using a wide range of topics and prompts, the results reveal that, indeed, these LLMs are linguistically different, particularly in terms of vocabulary, Part-Of-Speech (POS), and dependencies. In turn, these linguistic markers are shown to bring about a remarkable performance when applied to the LLM attribution task. 

\section{Methods and Materials}

\subsection{Data}
We build upon the highly influential Human ChatGPT Comparison Corpus (HC3) \cite{guo2023close}. HC3 is the most utilized database today for comparing LLM to human-generated texts. In its English portion, it encompasses both human and GPT-3.5 responses (termed ChatGPT in the original manuscript) to identical inputs divided into five distinct datasets: Finance \cite{maia201818}, Medicine \cite{medlog}, Long-Form Question Answering \cite{eli5} (aka reddit\_eli5), Open-Domain Question Answering (aka open\_qa) \cite{wikiqa},  and Computer Science (aka wiki\_csai dataset) \cite{guo2023close}. 
Here, we consider 1,000 randomly sampled inputs from each dataset, along with their GPT-3.5 responses. We then extend the database to include the matched responses provided by GPT-4 and Bard to these inputs. The resulting database, which we term the LLM Comparison Corpus (LC2), consists of 5,000 inputs and 15,000 responses (5,000 from each examined LLM). LC2, along with our entire code, is freely available here: \url{https://github.com/teddy4445/llm_style_diff}.

\subsection{Analytical Approach}

Drawing upon the comprehensive linguistic analysis of HC3 \cite{guo2023close}, we analyze the key linguistic features of LC2 as follows: we examine possible linguistic differences in terms of vocabulary, POS, dependencies, and sentiment across the three LLMs (see \cite{nlp} for a linguistic overview of these and related concepts). Statistically, for comparing vocabulary, we use an ANOVA test with Tukey post-hoc pairwise t-testing. Similarly, for comparing the distributions over POS and dependencies, we use Kolmagorov-Smirnov testing with the Bonferroni correction. For sentiment analysis, given its ordinal nature, we use a Wilcoxon signed-rank test.  Statistical significance is set to 0.05. Finally, we use the aforementioned linguistic markers as input to an off-the-shelf supervised machine-learning model (XGBoost \cite{xgboost}) for the task LLM attribution -- i.e., classifying a given text to its assumed LLM origin. The model's performance is reported in standard form. 

\section{Results}

\subsection{Vocabulary}

We start by examining the vocabulary presented by the studied LLMs. We focus on the following three characteristics: \textit{Average length} ($L$) -- the average number of words in each response; \textit{Vocabulary size} ($V$) --
the total number of unique words used in all responses; and \textit{Density} ($D$) -- which is defined as 
\[
D=\frac{100V}{L\cdot N}
\]
where N is the number of responses in the relevant dataset. The results are summarized in Table \ref{table:first}.
% across the five datasets, the LLMs' vocabulary varies significantly. 

Starting with the average length, the three LLMs are found to statistically differ from one another. Specifically, Bard provides the shortest average responses. In fact, it provides the shortest average answers in three out of the five datasets. 
% (open\_qa, reddit\_eli5, and wiki\_csai). 
In the remaining two datasets, Bard provides the longest average responses in one case and the second longest in the other.
In addition, GPT-3.5 provides longer average responses than GPT-4 (i.e., in four of the five datasets). 
Turning to vocabulary size, once more, Bard provides the smallest number of unique words in three out of the five datasets whereas in the remaining two, it comes second to GPT-4. Interestingly, GPT-4 provides greater vocabulary size than GPT-3.5 in four of the five settings. 
When considering density, GPT-4 demonstrates the highest density metric in three of the five datasets whereas in the remaining two datasets, Bard brings about a higher density. 

% Taken jointly, the three LLMs seem to exhibit slightly different communication strategies. 

Taken jointly, Bard seems to provide shorter responses compared to GPT-3.5 and GPT-4 while also presenting smaller vocabulary size and relatively high density. In addition, GPT-4 seems to provide shorter responses than GPT-3.5 while presenting higher vocabulary size and density. 
% Notably, differences are especially prominent in the open\_qa dataset where the average length of the provided responses by GPT3.5 is roughly two times longer than the alternatives, significantly 

\begin{table}[!ht]
\centering
\begin{tabular}{ccccc}
\hline \hline
\textbf{Dataset} & \textbf{LLM} &\textbf{ Average length }& \textbf{Vocabulary size} & \textbf{Density} \\ \hline \hline
\multirow{4}{*}{Finance} & GPT-3.5 & 208.13 & 20974 & 2.49 \\
 & GPT-4 & 197.53 & 22785 & 2.73 \\
 & Bard & 219.28 & 21809 & 2.64 \\ 
 \hline
\multirow{3}{*}{Medicine} & GPT-3.5 & 206.14 & 7910 & 3.11 \\
 & GPT-4 & 168.09 & 8827 & 5.69 \\
 & Bard & 180.16 & 7594 & 3.24 \\ \hline
\multirow{3}{*}{open\_qa} & GPT-3.5 & 142.61 & 15379 & 9.06 \\
 & GPT-4 & 88.42 & 12097 & 16.93 \\
 & Bard & 65.74 & 10829 & 17.34 \\ \hline
\multirow{3}{*}{reddit\_eli5} & GPT-3.5 & 191.38 & 45198 & 1.40 \\
 & GPT-4 & 151.18 & 48095 & 2.05 \\
 & Bard & 133.70 & 46147 & 1.87 \\ \hline
\multirow{3}{*}{wiki\_csai} & GPT-3.5 & 202.39 & 9347 & 5.03 \\
 & GPT-4 & 215.05 & 10074 & 6.73 \\
 & Bard & 186.18 & 9240 & 7.18 \\ \hline \hline
\end{tabular}
\caption{Average response length, vocabulary size, and density of GPT-3.5, GPT-4, and Bard using five datasets. }
\label{table:first}
\end{table}

\subsection{Part-of-Speech (POS)}

Next, we examine the POS distribution of the provided responses (the complete list, along with their abbreviations and complete results, are provided in the Appendix). The results are summarized in Figure \ref{fig:dist}. 
The three LLMs provide statistically different POS distributions. 

We start by examining the use of nouns and adjectives, the most prevalent POSs for all three LLMs. 
Starting with nouns, we see that GPT-3.5 uses nouns more often than the other LLMs both in singular (NN: 30.36\% vs 28.73\% by GPT-4 and 26.15\% by Bard) and plural form (NNS: 16.19\% vs 13.51\% by GPT-4 and 12.67\% by Bard). In turn, GPT-4 uses nouns more often than Bard in both forms, although the differences are much smaller.
That is not the case for singular-form proper nouns (NNP), which are around twice less prevalent than non-proper nouns (NN), where GPT-4 demonstrates higher use patterns than Bard (11.43\% vs 9.91\%) which, in turn, uses them more often than GPT-3.5 (8.55\%). When plural-form proper nouns (NNPS) are considered, all LLMs scarcely use them (all are below 0.3\% prevalence). However, GPT-3.5 uses them almost two times less often than the others (0.17\% compared to 0.29\% and 0.26\% by GPT-4 and Bard, respectively).
Turning to adjectives (JJ) and the significantly less prevalent comparative adjectives (JJR), once more, we see that GPT-3.5 provides the greatest prevalence followed by GPT-4 and Bard (JJ: 15.12\%, 14.15\%, 12.75\%; JJR: 0.72\%, 0.68\%, 0.89\%). When superlative adjectives (JJS) are concerned, all LLMs rarely use them (all are below 0.9\%). However, Bard uses them more often as the GPT-4 which, in turn, uses them more than GPT-3.5 (0.82\% vs 0.26\%, 0.19\%).

Considering verbs and adverbs, the results point to very minor differences between the LLMs. Specifically, GPT-3.5 slightly uses present tense verbs (VBG and VBP) more often than the others whereas Bard uses past verbs (VBN and VBD) and third-person verbs (VBZ) slightly more often. In turn, GPT-4 presents a marginally higher use of base-form verbs (VB) than the others (for the complete results, see Appendix). Turning to adverbs, GPT-3.5 presents a lower use of adverbs (3.66\%) compared to GPT-4 and Bard (4.87\% and 4.15\%, respectively). 

The remaining POSs (7.26\%-23\% of the text), encompass a relatively small portion of the text. However, considerable differences between the LLMs are encountered. Specifically, Bard presents a relatively high use of personal pronouns (PRP: 1.7\% vs 0.69\% by GPT-3.5 and 0.7\% by GPT-4), existential theres (EX: 0.92\% vs 0.13\% by GPT-3.5 and 0.09\% by GPT-4),  coordinating conjunctions (CC: 0.76\% vs 0.02\% by GPT-4 and 0.11\% by GPT-3.5), foreign words (FW: 0.45\% vs 0.03\% by GPT-3.5 and 0.05\% by GPT-4), list markers (LS: 0.55\% vs 0\% by GPT-3.5 and 0\% by GPT-4), wh-pronouns (WP: 0.58\% vs 0\% by GPT-3.5 and 0.01\% by GPT-4), wh-determiners (WDT: 1.00\% vs 0\% by GPT-3.5 and 0.01\% by GPT-4), and interjections (UH: 0.53\% vs 0.06\% by GPT-3.5 and 0.08\% by GPT-4). More broadly speaking, when non-negligible use of POSs is considered (i.e., over 0.25\% prevalence), Bard presents conspicuously diverse use patterns (88.5\%) compared to GPT-3.5 and GPT-4 (48.6\% and 60.0\%, respectively).

\begin{figure}
    \centering
    \includegraphics[width=0.99\textwidth]{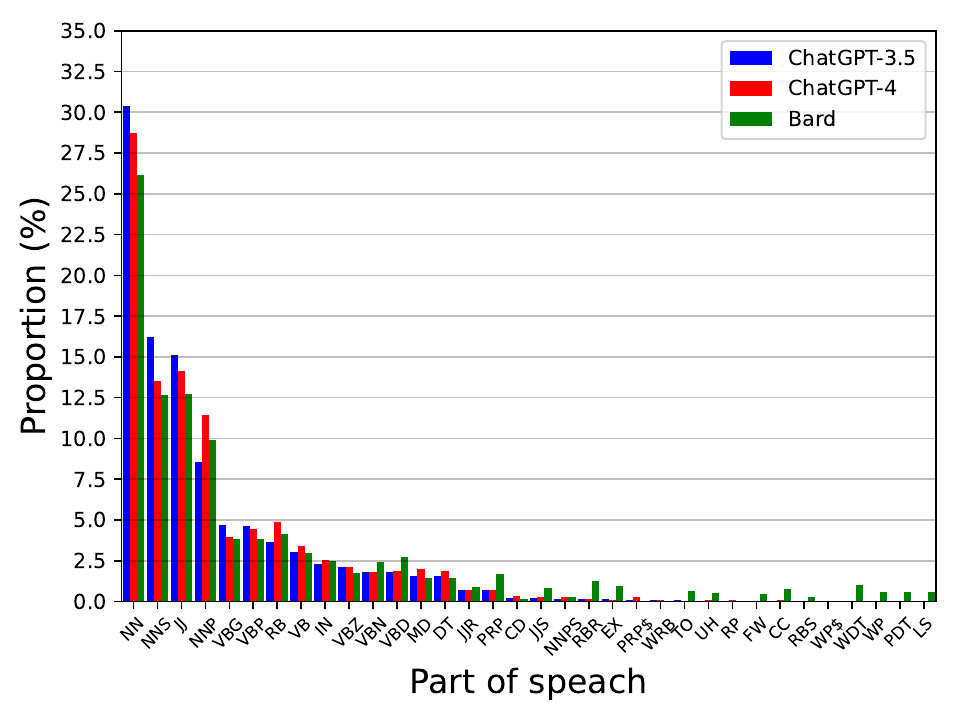}
    \caption{Part-of-Speech (POS) distribution comparison between GPT-3.5, GPT-4, and Bard.}
    \label{fig:dist}
\end{figure}

Overall, notable differences are observed in the POS distributions associated with the examined LLMs. The differences seem to span over both highly frequent POSs (i.e., noun) and low-frequency POSs (i.e., existential there). Most notably, significant differences are observed in the presence of low-frequency POSs with Bard typically using them more often.

\subsection{Dependency Parsing}

Next, we examine the distribution of key dependencies in the provided responses (the complete list, along with their abbreviations and complete results, are provided in the Appendix). The results are summarized in Figure \ref{fig:fi}. 
Statistically, GPT-3.5 and GPT-4 do not differ significantly in terms of dependencies ($p=0.13$) while BARD does differ from both at \(p < 0.05\). 

\begin{figure}
    \centering
    \includegraphics[width=0.99\textwidth]{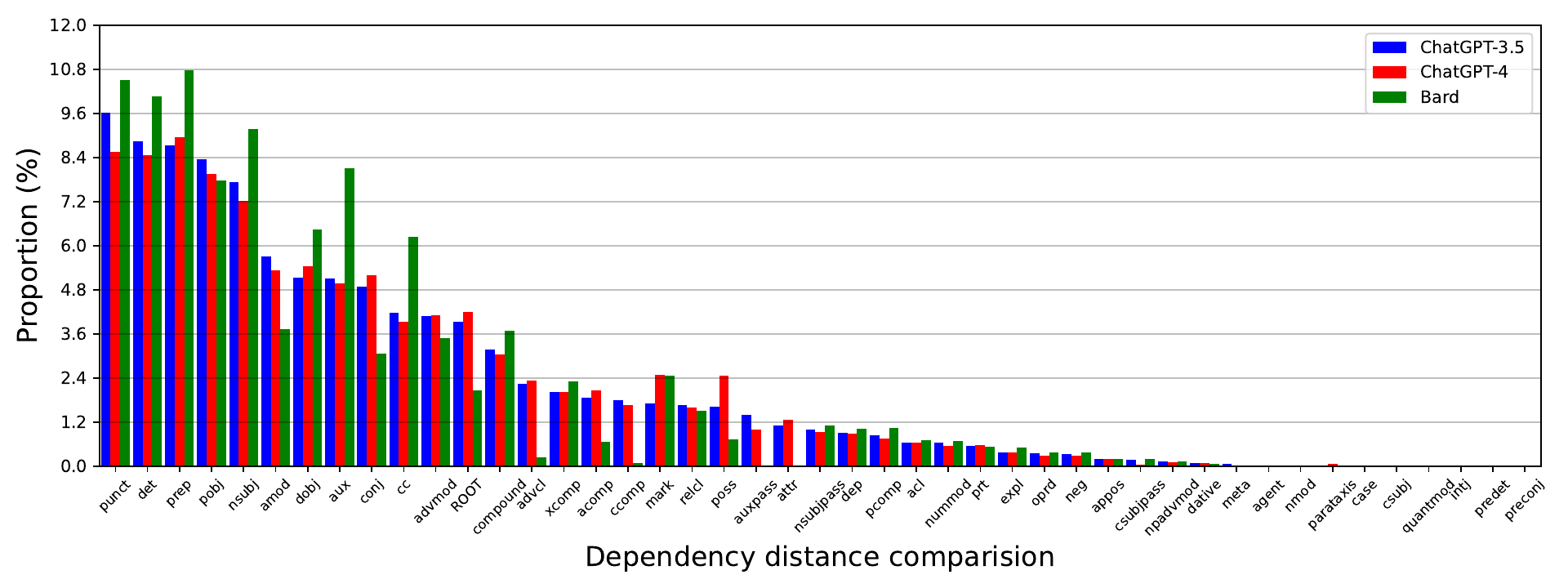}
    \caption{Top-30 dependency relations comparison between GPT-3.5, GPT-4, and Bard.}
    \label{fig:sentiment}
\end{figure}

% Following the Universal Stanford Dependencies taxonomy \cite{de2014universal}, we discuss the key types below.
% (de Marneffe and Manning, 2008): Stanford typed dependencies manual
Given the wide variety of dependency types, we discuss the prominent differences only for dependencies with at least 1\% prevalence (for at least one of the examined LLMs). 
Starting with the three most prevalent dependency types in the data, punctuation, determiner, and preposition, we see that Bard uses them more often than GPT-3.5 and GPT-4 (punct: 10.5\% vs 9.62\%, 8.56\%; det: 10.06\% vs 8.84\%, 8.46\%; prep: 10.78\% vs 8.73\%, 8.95\%). Other noteworthy differences include Bard's higher use of nominal subjects (nsubj: 9.17\% vs 7.72\%, 7.23\%), direct objects (dobj: 6.44\% vs 5.13\%, 5.44\%), auxiliary (aux: 8.12\% vs 5.12\%, 4.96\%) and coordinating conjunctions (cc: 6.23\% vs 4.18\%, 3.93\%),
% and prepositional complement (pcomp: 1.04\% vs 0.84\%, 0.76\%),
and lower use of adjectival modifier (amod: 3.72\% vs 5.13\%, 5.44\%), conjunct (conj: 3.07\% vs 4.87\%, 5.2\%), root dependencies (ROOT: 2.06\% vs 3.92\%, 4.19\%), adverbial clause modifier (advcl: 0.23\% vs 2.24\% and 2.33\%), adjectival complement (acomp: 0.67\% vs 1.85\%, 2.07\%), clausal complement (ccomp: 0.08\% vs  1.79\%, 1.66\%), possession modifiers (poss: 0.72\% vs 1.62\%, 2.45\%) , passive auxiliaries (auxpass: 0\% vs 1.39\%, 0.99\%) and attributions (attr: 0\% vs 1.11\%, 1.27\%).

Overall, while no significant differences are observed between GPT-3.5 and GPT-4, Bard seems to significantly differ from both with a distinctively higher use of highly frequent dependencies (i.e., punctuation, determination, preposition, direct object, auxiliary verb, etc) and a lower use of other dependencies (i.e., adverbial clause modifier, adjectival complement, clausal complement, attributions, etc).  

\subsection{Sentiment Analysis}

Considering sentiment, the three LLMs do not present statistically significant differences \(p=0.284\). All three LLMs tend to produce positive sentiment (GPT-3.5: 53.8\%; GPT-4: 53.3\%; and Bard: 53.22\%) with relatively few cases of neutral sentiment (GPT-3.5: 2.96\%; GPT-4: 3.75\%; and Bard: 2.89\%). 

\begin{figure}
    \centering
    \includegraphics[width=0.99\textwidth]{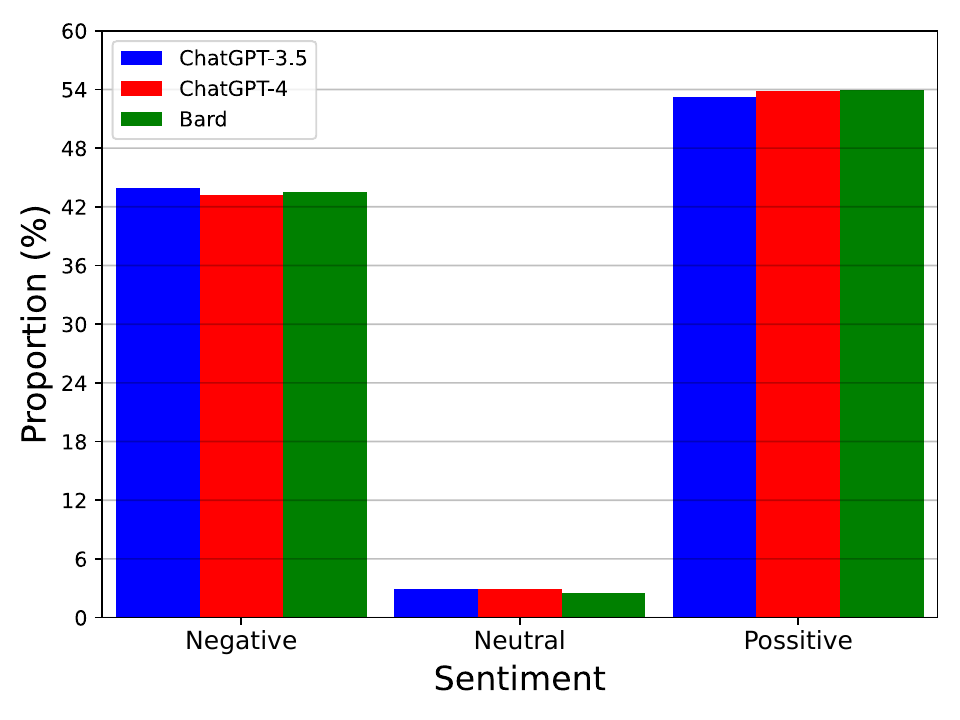}
    \caption{Sentiment distribution over the responses generated by GPT-3.5, GPT-4, and Bard.}
    \label{fig:sentiment}
\end{figure}

\subsection{LLM Attribution}
Here, we consider the task of LLM attribution. That is, given a text, we wish to classify it to its LLM source. Recall that in this study, we are primarily interested in investigating the linguistic differences between LLMs and their potential use as indicators for LLM attribution. As such, we opt for an off-the-shelf XGboost model that receives the above linguistic profile of a given text as input (i.e., vocabulary, POS distribution, dependency distribution, and sentiment) and classify it to one of the three LLMs (i.e., its origin). We used a k-fold approach with \(k=5\) such that for each fold, the portion of the train and test instances from each of the five datasets are identical. Note that, presumably, more advanced and fine-tuned models could be developed for the task of LLM attribution, however, these are outside the focus of our analysis. 

Our results show a high accuracy of \(0.88\) with \(F_1\) score of \(0.87\). Table \ref{table:classification_performance} presents the classification performance of the XGboost model for each LLM and the average performance. The results are shown as the mean of \(k=5\) folds.

\begin{figure}
    \centering
    \includegraphics[width=0.99\textwidth]{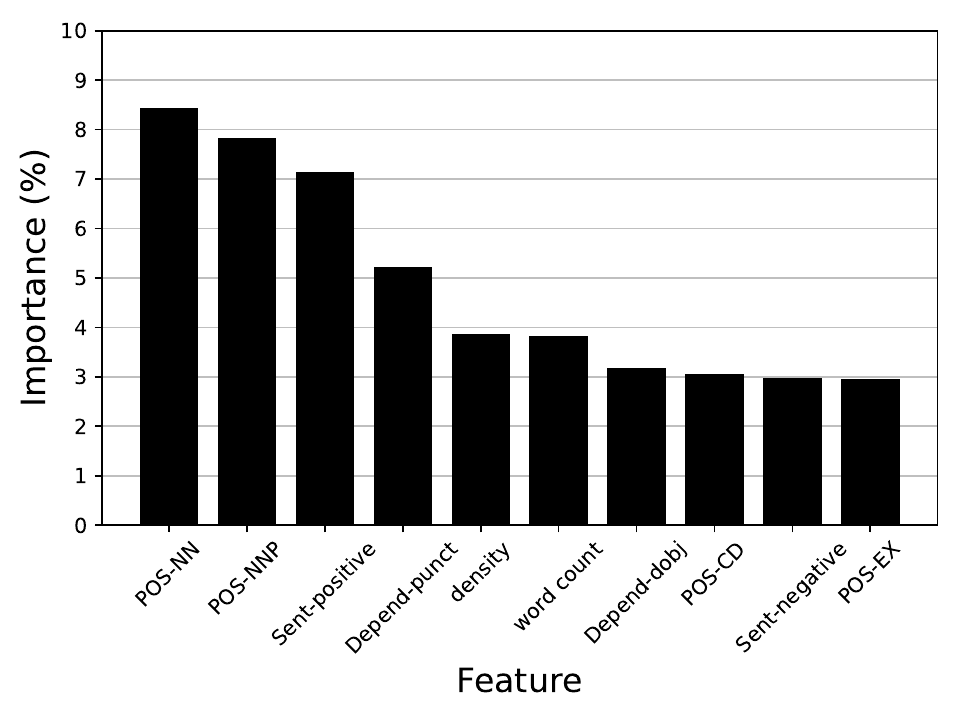}
    \caption{Feature importance of the top-10 most important features of the XGboost model.}
    \label{fig:fi}
\end{figure}

Feature importance is extracted using the information gain method and is reported in Figure \ref{fig:fi}. Note that linguistic features from all examined aspects (i.e., vocabulary, POS, dependency, and sentiment) are present in the top ten most important features of the model. These are led by the noun (NN) and proper noun (NNP) POSs, positive sentiment, punctuation dependency (punct), and the vocabulary's density and word count. 

\begin{table}[!ht]
\centering
\begin{tabular}{lccccc}
\hline \hline
\textbf{} & \textbf{Prediction} & \textbf{Recall} & \textbf{\(F_1\) score} & \textbf{Accuracy} & \textbf{Support} \\ \hline \hline
\textbf{GPT-3.5} & 0.90 & 0.82 & 0.86 & 0.87 & 1000 \\
\textbf{GPT-4} & 0.82 & 0.84 & 0.83 & 0.85 & 1000 \\
\textbf{Bard} & 0.94 & 0.89 & 0.91 & 0.92 & 1000 \\ \hline
\textbf{Average} & 0.89 & 0.85 & 0.87 & 0.88 & 3000 \\ \hline \hline
\end{tabular}
\caption{Classification performance of the linguistic-based XGboost model for each LLM and the average performance. The results are shown as the mean of \(k=5\) folds.}
\label{table:classification_performance}
\end{table}

\section{Discussion}

In this study, we analyzed and compared the linguistic characteristics of three of the most popular LLMS today: GPT-3.5, GPT-4, and Bard. Our results combine to suggest that, just like human authors, LLMs are linguistically distinguishable. Specifically, the results show that different LLMs present different linguistic markers, especially in terms of vocabulary, POS, and dependencies. These and similar differences are shown to be effectively used for LLM attribution - i.e., the identification of which LLM generated a given text, with a remarkably high accuracy. 

This key result has important theoretical and practical implications. 
From a theoretical perspective, this result seems to affirm the potential of LLMs to reproduce some of the diversity inherent in human language expression. That is, different LLMs tend to display different linguistic styles, akin to how human authors do. Furthermore, the observed differences could advance our understanding of the potential impact of various design and training decisions on the linguistic style presented by an LLM. We plan to conduct such an inquiry in future work. From a practical perspective, the identified differences could aid in model comparison and evaluation, and guide the selection and development of LLMs for specific purposes and applications \cite{ghosal2023survey,verma2023ghostbuster,macko2024authorship}. Moreover, we believe that the results should be taken into consideration for the important task of LLM detection \cite{end_1,end_2}. Specifically, the detection of LLM-generated texts, or even LLM-assisted writing, could potentially be improved by considering LLM-specific linguistic markers and style as detectors trained with texts generated by one LLM are known to face difficulties when challenged with texts generated by another \cite{wang2023m4}. 

This study is not without limitations. First, our analysis focuses on a diverse, yet limited, cohort of datasets. As such, an investigation into different datasets may result in slightly different outcomes, depending on the nature of the inputs provided to the LLM. Second, the linguistic style is collected for the zero-shot case where the LLMs do not have any context. Thus, our study focuses on the most simplistic use-case of these advanced tools where context does not play a role. Future works may replicate our analysis with different datasets and contexts, providing a wider perspective on the matter. Additionally, our analysis focuses on the English language. Studying the differences between LLMs across languages can be valuable to a wider range of applications and use cases. 

\bibliographystyle{acm}
\bibliography{bib}

\section*{Appendix}

\begin{table}[!ht]
\centering
\begin{tabular}{|l|c|c|c|}
\hline 
\textbf{LLM} & \textbf{Negative} & \textbf{Neutral} & \textbf{Positive} \\ \hline 
GPT-3.5 & 43.24 & 2.96 & 53.80 \\ \hline 
GPT-4  & 42.95 & 3.75 & 53.30 \\ \hline 
Bard & 43.89 & 2.89 & 53.22 \\ \hline 
\end{tabular}
\caption{Sentiment distribution.}
\label{table:sentiment}
\end{table}

\begin{table}[!ht]
    \centering
    \begin{tabular}{|l|l|l|l|l|}
    \hline
        \textbf{POS} & \textbf{Symbol} & \textbf{GPT-3.5} & \textbf{GPT-4} & \textbf{Bard} \\ \hline
       noun singular &  NN & 30.36 & 28.73 & 26.15 \\ \hline
      noun plural   &  NNS & 16.19 & 13.51 & 12.67 \\ \hline
      adjective  &  JJ & 15.12 & 14.15 & 12.75 \\ \hline
      proper noun, singular   & NNP & 8.55 & 11.43 & 9.91 \\ \hline
       verb, gerund/present participle   & VBG & 4.66 & 3.93 & 3.84 \\ \hline
      verb, singular, present, non-3d &   VBP & 4.61 & 4.45 & 3.83 \\ \hline
      adverb    &  RB & 3.66 & 4.87 & 4.15 \\ \hline
       verb, base form   &  VB & 3.04 & 3.42 & 2.96 \\ \hline
      preposition/subordinating conjunction  &  IN & 2.27 & 2.52 & 2.49 \\ \hline
      verb, 3rd person singular present  &  VBZ & 2.12 & 2.11 & 1.77 \\ \hline
     verb, past participle   &  VBN & 1.8 & 1.81 & 2.43 \\ \hline
       verb, past tense  &  VBD & 1.79 & 1.88 & 2.74 \\ \hline
       modal &  MD & 1.58 & 1.99 & 1.45 \\ \hline
      determiner  &  DT & 1.55 & 1.9 & 1.45 \\ \hline
     adjective, comparative   &  JJR & 0.72 & 0.68 & 0.89 \\ \hline
     personal pronoun  &   PRP & 0.69 & 0.7 & 1.7 \\ \hline
    cardinal digit   &   CD & 0.21 & 0.35 & 0.18 \\ \hline
     adjective, superlative   &  JJS & 0.19 & 0.26 & 0.82 \\ \hline
     proper noun, plural   &   NNPS & 0.17 & 0.29 & 0.26 \\ \hline
     adverb, comparative   &  RBR & 0.15 & 0.17 & 1.25 \\ \hline
     existential there   &  EX & 0.13 & 0.09 & 0.92 \\ \hline
     possessive pronoun  &   PRP\$ & 0.1 & 0.28 & 0 \\ \hline
      wh-abverb &  WRB & 0.1 & 0.08 & 0 \\ \hline
       to go ‘to’ &   TO & 0.08 & 0.05 & 0.65 \\ \hline
     interjection   &   UH & 0.06 & 0.08 & 0.53 \\ \hline
      particle   &  RP & 0.04 & 0.06 & 0 \\ \hline
      foreign word   &  FW & 0.03 & 0.05 & 0.45 \\ \hline
       coordinating conjunction  &  CC & 0.02 & 0.11 & 0.76 \\ \hline
      possessive ending  &   RBS & 0.02 & 0.01 & 0.27 \\ \hline
    possessive wh-pronoun   &   WP\$ & 0 & 0.01 & 0 \\ \hline
     wh-determiner    &  WDT & 0 & 0.01 & 1 \\ \hline
      wh-pronoun   &  WP & 0 & 0.01 & 0.58 \\ \hline
      predeterminer &   PDT & 0 & 0 & 0.59 \\ \hline
      list marker &   LS & 0 & 0 & 0.55 \\ \hline
    \end{tabular}
\end{table}

\begin{table}[!ht]
    \centering
    \begin{tabular}{|l|l|l|l|l|}
    \hline
        \textbf{Dependency} & \textbf{Symbol} & \textbf{GPT-3.5} & \textbf{GPT-4} & \textbf{Bard} \\ \hline
        punctuation & punct & 9.62 & 8.56 & 10.5 \\ \hline
       determiner  & det & 8.84 & 8.46 & 10.06 \\ \hline
      Prepositional modifier &   prep & 8.73 & 8.95 & 10.78 \\ \hline
       Preposition object&   pobj & 8.36 & 7.95 & 7.77 \\ \hline
       Nominal subject &  nsubj & 7.72 & 7.23 & 9.17 \\ \hline
      Adjectival modifier &    amod & 5.72 & 5.33 & 3.72 \\ \hline
      Direct Object &   dobj & 5.13 & 5.44 & 6.44 \\ \hline
      Auxiliary verb  &  aux & 5.12 & 4.96 & 8.12 \\ \hline
      Conjunct &   conj & 4.87 & 5.2 & 3.07 \\ \hline
      Coordinating conjunction &   cc & 4.18 & 3.93 & 6.23 \\ \hline
      Adverbial modifier &   advmod & 4.08 & 4.1 & 3.49 \\ \hline
     Root  &   ROOT & 3.92 & 4.19 & 2.06 \\ \hline
       Compound word &   compound & 3.17 & 3.05 & 3.69 \\ \hline
      Adverbial clause modifier &   advcl & 2.24 & 2.33 & 0.23 \\ \hline
     Open clausal complement  &   xcomp & 2.01 & 2.02 & 2.3 \\ \hline
      Adjectival complement &   acomp & 1.85 & 2.07 & 0.67 \\ \hline
       Clausal complement &   ccomp & 1.79 & 1.66 & 0.08 \\ \hline
       Marker &  mark & 1.71 & 2.47 & 2.47 \\ \hline
       Relative clause modifier &  relcl & 1.66 & 1.6 & 1.5 \\ \hline
        Possession modifier &  poss & 1.62 & 2.45 & 0.72 \\ \hline
        Auxiliary verb (passive)	 &  auxpass & 1.39 & 0.99 & 0 \\ \hline
       Attribute &  attr & 1.11 & 1.27 & 0 \\ \hline
      Nominal subject (passive)  &  nsubjpass & 0.99 & 0.93 & 1.1 \\ \hline
      Unclassified dependent &  dep & 0.9 & 0.88 & 1.02 \\ \hline
       Preposition complement &  pcomp & 0.84 & 0.76 & 1.04 \\ \hline
       Clausal modifier of noun &  acl & 0.64 & 0.64 & 0.71 \\ \hline
       Number &   nummod & 0.64 & 0.56 & 0.68 \\ \hline
       Verb particle &  prt & 0.54 & 0.57 & 0.54 \\ \hline
       Expletive &  expl & 0.37 & 0.37 & 0.5 \\ \hline
       Object predicate &  oprd & 0.34 & 0.29 & 0.37 \\ \hline
        Negation modifier &  neg & 0.32 & 0.29 & 0.38 \\ \hline
       Appositional modifier &  appos & 0.19 & 0.19 & 0.2 \\ \hline
        Clausal subject (passive) &  csubjpass & 0.19 & 0.05 & 0.19 \\ \hline
       Noun phrase as adverbial modifier &  npadvmod & 0.13 & 0.1 & 0.14 \\ \hline
       Dative &  dative & 0.08 & 0.08 & 0.06 \\ \hline
       Meta data &   meta & 0.05 & 0.01 & 0 \\ \hline
        Agent (passive) &  agent & 0 & 0 & 0 \\ \hline
       Modifier of nominal  &  nmod & 0 & 0 & 0 \\ \hline
       Parataxis &   parataxis & 0 & 0.06 & 0 \\ \hline
      Case marker &   case & 0 & 0 & 0 \\ \hline
       Clausal subject &   csubj & 0 & 0.01 & 0 \\ \hline
    \end{tabular}
\end{table}

\end{document}